\documentclass[letterpaper]{article} 
\usepackage{aaai24}  
\usepackage{times}  
\usepackage{helvet}  
\usepackage{courier}  
\usepackage[hyphens]{url}  
\usepackage{graphicx} 
\usepackage{natbib}  
\usepackage{caption} 
\usepackage{algorithm}
\usepackage{algorithmic}

\usepackage{graphicx}
\usepackage{amsmath}
\usepackage{amssymb}
\usepackage{booktabs}
\usepackage{bbding}
\usepackage{color}
\usepackage[T1]{fontenc}
\usepackage{multirow}

\newtheorem{defi}{Definition}

\newtheorem{prop}{Proposition}
\newtheorem{proof}{Proof}
\usepackage[switch]{lineno}

\usepackage{newfloat}
\usepackage{listings}
\DeclareCaptionStyle{ruled}{labelfont=normalfont,labelsep=colon,strut=off} 
\floatstyle{ruled}
\newfloat{listing}{tb}{lst}{}
\floatname{listing}{Listing}
\title{MCA: Moment Channel Attention Networks}
\author {
    Yangbo Jiang\textsuperscript{\rm 1,2},
    Zhiwei Jiang\textsuperscript{\rm 3}, 
    Le Han\textsuperscript{\rm 1,2},
    Zenan Huang\textsuperscript{\rm 1,2},
    Nenggan Zheng\textsuperscript{\rm 1,2,4,5}\thanks{Corresponding author}
}
\affiliations {
    \textsuperscript{\rm 1}Qiushi Academy for Advanced Studies, Zhejiang University, Hangzhou, Zhejiang, China\\ 
    \textsuperscript{\rm 2}College of Computer Science and Technology, Zhejiang University, Hangzhou, Zhejiang, China\\
    \textsuperscript{\rm 3}Guangzhou Electronic Technology Co., Ltd., Chinese Academy of Sciences, GuangZhou, China\\
    \textsuperscript{\rm 4}State Key Lab of Brain-Machine Intelligence, Zhejiang University, Hangzhou, Zhejiang, China\\
    \textsuperscript{\rm 5}CCAI by MOE and Zhejiang Provincial Government(ZJU), Hangzhou, Zhejiang, China\\
    \{jiangyangbo, hanle, lccurious\}@zju.edu.cn, j\_z\_w@163.com, zng@cs.zju.edu.cn
 
}

\usepackage{bibentry}

\begin{document}

\maketitle

\begin{abstract}
    Channel attention mechanisms endeavor to recalibrate channel weights to enhance representation abilities of networks. However, mainstream methods often rely solely on global average pooling as the feature squeezer, which significantly limits the overall potential of models.
    In this paper, we investigate the statistical moments of feature maps within a neural network. Our findings highlight the critical role of high-order moments in enhancing model capacity.
    Consequently, we introduce a flexible and comprehensive mechanism termed Extensive Moment Aggregation (EMA) to capture the global spatial context.
    Building upon this mechanism, we propose the Moment Channel Attention (MCA) framework, which efficiently incorporates multiple levels of moment-based information while minimizing additional computation costs through our Cross Moment Convolution (CMC) module. The CMC module via channel-wise convolution layer to capture multiple order moment information as well as cross channel features.
    The MCA block is designed to be lightweight and easily integrated into a variety of neural network architectures. 
    Experimental results on classical image classification, object detection, and instance segmentation tasks demonstrate that our proposed method achieves state-of-the-art results, outperforming existing channel attention methods. 
  
\end{abstract}
  
\section{Introduction}

Channel attention mechanisms have aroused wide concern in the field of computer vision due to their remarkable performance on various tasks, including image classification\cite{hu2018squeeze,woo2018cbam}, object detection\cite{dai2017deformable,carion2020end}, instance segmentation\cite{yuan2018ocnet,fu2019dual}, face recognition\cite{yang2017neural,wang2020hierarchical}, image generation \cite{gregor2015draw,zhang2019self,liu2022using} and multi-modal learning \cite{su2019vl,xu2018attngan}, etc. 
The objective of channel attention method is to learn the aggregation of global spatial information and adaptively recalibrate the weight of each channel. As an effective method, the channel attention approach has become an essential module in various neural networks and has been widely employed to enhance the representational power of network. 

\begin{figure}[ht]
	\centering
	\includegraphics[width=8.5 cm]{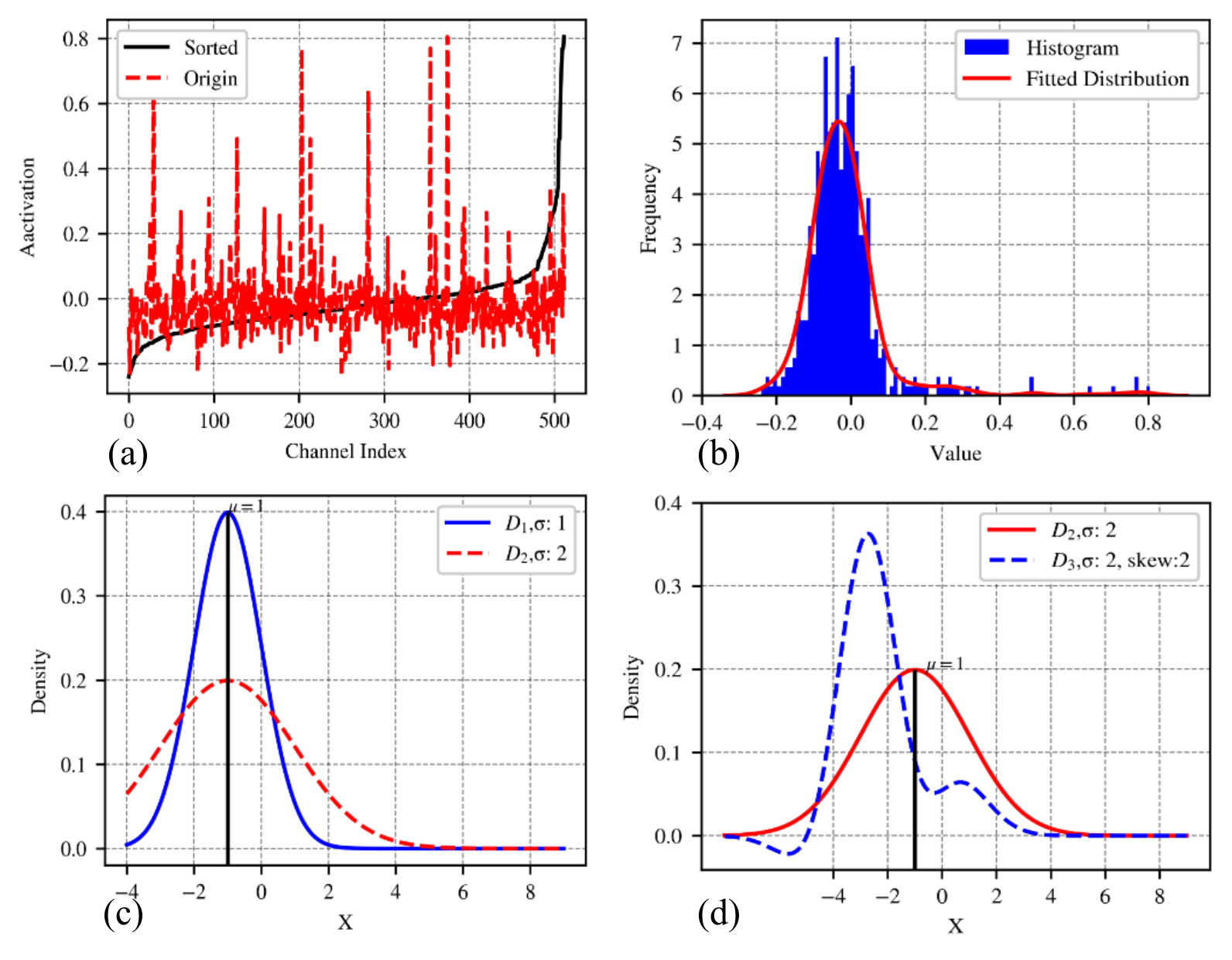}  
	\caption{The activation of feature map exhibits a distinct probability distribution, as illustrated in (a) and (b). While the first-order moment is inadequate to represent a standard Gaussian distribution in (c), and combining the first and second-order moment falls short in capturing non-Gaussian distributions (as depicted in (d)). Extensive moment aggregation mechanism offers a viable solution to this challenge. }
	\label{fig:fig1}
\end{figure}

\begin{table*}[]
	\small
	\centering
\setlength{\tabcolsep}{2pt}
		\begin{tabular}{l|cc|ccc|c|c}
			\hline
			\multicolumn{1}{c|}{\multirow{2}{*}{Blocks}} & \multicolumn{2}{c|}{Squeeze Method} & \multicolumn{3}{c|}{Excitation  Method}       & \multirow{2}{*}{Params} & \multirow{2}{*}{AP}  \\ \cline{2-6}  
			\multicolumn{1}{c|}{}      & \multicolumn{1}{c|}{Pooling Method}  & \multicolumn{1}{c|}{Order} & \multicolumn{1}{c|}{Method} & \multicolumn{1}{c|}{ Moments  Fusion  } & \multicolumn{1}{c|}{ Channel  Interaction  }   &     &                \\ \hline
			SE\quad (CVPR,2018)     & \multicolumn{1}{c|}{GAP}      &   \multicolumn{1}{|c|} {1ST }    &  \multicolumn{1}{c|}{ FC }   &     \multicolumn{1}{c|}{\XSolidBrush}  &     \multicolumn{1}{c|}{\XSolidBrush}    & $\frac{2}{r} \sum_{s=1}^{S}N_{s}\cdot C_{s}^{2}  $  & 35.4    \\ 
			SRM (ICCV,2019)   & \multicolumn{1}{c|}{GAP \& STD}       &    \multicolumn{1}{|c|}{1ST \& 2ND }     &   \multicolumn{1}{c|}{CFC}        &  \multicolumn{1}{c|}{\Checkmark }     &  \multicolumn{1}{c|}{\XSolidBrush }  & $\sum_{s=1}^{S}N_{s}\cdot C_{s} *6 $ & 36.9   \\ 
			ECA (CVPR,2020)    & \multicolumn{1}{c|}{GAP}      &   \multicolumn{1}{|c|} {1ST  }  &   \multicolumn{1}{c|}{   Conv1D}   &     \multicolumn{1}{c|}{\XSolidBrush }  &   \multicolumn{1}{c|}{\Checkmark }    & $ \sum_{s=1}^{S}N_{s}\cdot |( log_{2}C +1) /2 |_{odd}$  & 36.2                \\ 
			GCT (CVPR,2021)   & \multicolumn{1}{c|}{GAP}       &    \multicolumn{1}{|c|}{1ST }     &   \multicolumn{1}{c|}{  Gaussian}  &   \multicolumn{1}{c|}{\XSolidBrush }      &  \multicolumn{1}{c|}{\Checkmark }     & $\sum_{s=1}^{S}N_{s} $ & 37.9   \\ 
			MCA-E (Ours)    & \multicolumn{1}{c|}{GAP \& VAR} &  \multicolumn{1}{|c|} {1ST \& 2ND }  &  \multicolumn{1}{c|}{ CMC}     &   \multicolumn{1}{c|}{\Checkmark }    &\multicolumn{1}{c|}{\Checkmark}    & $\sum_{s=1}^{S}N_{s}\cdot C_{s} *2$ & 38.0          \\ 
			MCA-S (Ours)   & \multicolumn{1}{c|}{GAP \& SKEW}&  \multicolumn{1}{|c|} {1ST \& 3RD  }  &     \multicolumn{1}{c|}{CMC} &  \multicolumn{1}{c|}{\Checkmark }   &   \multicolumn{1}{c|}{\Checkmark}      & $\sum_{s=1}^{S}N_{s}\cdot C_{s} *2$ & \textbf{38.3}   \\ \hline
			
		\end{tabular}
	\caption{ Comparison with other SOTA channel attention blocks on COCO val2017 set. Our proposed block is distinguished by two key features: (i) a flexible and comprehensive moment aggregation approach for pooling; (ii) an novel excitation method that serves two efficiencies: moment fusion and local channel interaction. Here, FC, CFC, and CMC refer to fully connected, channel-wise fully connected, and cross moment conv1D layer, respectively. While Gaussian denotes Gaussian function. $C$ denotes the number of channels. $r$ denotes the reduction ratio of SE. $\left | \cdot \right | $ odd indicates the nearest odd number of $\cdot$. }
	\label{tbl:table1}
\end{table*}

In recent years, several effective channel attention methods have been presented. Hu et al. \cite{hu2018squeeze} introduced the Squeeze-and-Excitation Networks (SENet) with squeeze and excitation modules. The squeeze module maps global spatial information with global average pooling (GAP), while the excitation module recalibrates the weights of each channel based on their channel-wise relationships. Since then, other channel attention algorithms \cite{lee2019srm,Wang2020ECANetEC,ruan2021gaussian} have achieved significant improvements to both the squeeze and excitation modules.
From a statistical perspective, the squeeze module of SENet \cite{hu2018squeeze}, ECANet \cite{Wang2020ECANetEC}, and GCT \cite{ruan2021gaussian} utilize first-order statistics (GAP) to extract global information from the feature map without considering high-order statistics. 
Despite SRM \cite{lee2019srm} involving the utilization of both first-order statistics (GAP or Mean) and second-order statistics (Standard Deviation, STD) for spatial representation, it merely concatenates them with a basic multiply and addition operation. However, the local channel interaction is absent in the methodology of SRM.

To dig deeper into the shortcomings of the squeeze module, we investigate the moment statistics within the probability distribution. The activation of feature map within a neural network can be regarded as samples from a probability distribution, as depicted in Fig.~\ref{fig:fig1}(a) and (b). 
Moments provide a method to represent this probability distribution by employing the moment generating function \cite{casella2002statistical}. These moments are derived from the Taylor series expansion of the expectation of a random variable \cite{athreya2006measure,garthwaite2002statistical}. Specifically, for a Gaussian distribution, the first-order moment corresponds to the Mean ($\mu$), and the second-order moment corresponds to the Variance (VAR, $\sigma^{2}$), and the higher order moments is zero or constant \cite{patel1996handbook,bai2005tests}. 
From a statistical perspective, as shown in the moment generate function, other order moment, except for the first-order moment, second and third-order moment pay a crucial role in represent a probability distribution.
For example, in Fig.~\ref{fig:fig1}(c), the probability distribution $D_{1}$ and $D_{2}$ both have the same mean of -1, but the former has a lower STD of 1 compared to the latter's 2. A smaller STD indicates a more concentrated distance from the Mean, while the Mean reflects the degree of concentration of the data. Similarly,  in Fig.~\ref{fig:fig1}(d), the $D_{2}$, being a Gaussian distribution, has a lower SKEW (third-order moment) of 0 compared to the $D_{3}$'s 2, and the SKEW indicates the asymmetry of the distribution. 
Therefore, a flexible multiple moment aggregation method is crucial for effectively representing the feature map of networks.

In this paper, we propose a Moment Channel Attention (MCA) method inspired by moment representation in the probability distribution. 
The MCA method employs Extensive Moment Aggregation (EMA) mechanism as the pooling method for a refined aggregation of extensive moment information. Furthermore, we introduce the Cross Moment Convolution (CMC) method via channel-wise convolution layer to capture the interaction between moments of different orders in the feature map, as well as local cross channels feature.  
Based on EMA mechanism with base term and other terms, we present two MCA methods: MCA-E and MCA-S. As shown in Table~\ref{tbl:table1}, MCA-E and MCA-S slightly increase  the model size. Despite this slight increase in computational cost, the accuracy of the model is effectively improved. MCA-E and MCA-S achieve excellent performance compared to other SOTA methods. 
Moreover, MCA method is easily integrated into network architectures, such as ResNets \cite{he2016deep,xie2017aggregated}, and lighter architectures like ShuffleNet \cite{ma2018shufflenet}, Inception \cite{ioffe2015batch}. 
Furthermore, MCA can be trained end-to-end, making it convenient and flexible for practical applications.
In general, the main contributions of this work are summarized as three-fold: 
\begin{itemize}
\item  

In this paper, We regard the activation of feature map as samples following a probability distribution and analyzing the validity of each order moment. 
Consequently, we introduce the EMA mechanism, revealing the effectiveness of utilizing a constrained set of moments to accurately represent the global spatial features.
Building upon this aggregation strategy, we propose the MCA method, which captures feature from multiple moments. Notably, the ECA method can be viewed as a specific instance of our proposed method.

\item Furthermore, we introduce the CMC method to moment fusion. The CMC method employs a concise and powerful channel-wise convolution layer that merges information from multiple order moments within the feature map, while concurrently incorporating local cross-channel interactions.

\item  Compared with existing  channel attention methods, our proposed method achieves excellent state-of-the-art results on the COCO benchmarks for object detection and instance segment, as well as the ImageNet classification benchmark. In addition, our method is a lightweight and convenient, with computational and model complexity comparable to ECA  method.
\end{itemize}

\section{Related Work}
\subsubsection{Attention Mechanisms} 
Attention mechanism is an effective way to enhance the representational power of neural networks, which can be typically categorized into channel attention \cite{hu2018squeeze,gao2019global,lee2019srm,Wang2020ECANetEC, yang2020gated, qin2021fcanet,ruan2021gaussian}, spatial attention \cite{wang2018non,chen20182,cao2019gcnet,li2019selective}, and channel \& spatial attention \cite{park2018bam,woo2018cbam,roy2018recalibrating}. We briefly review channel attention methods firstly, which aims to recalibrate the weights of channels to facilitate the  representation ability.
SENet \cite{hu2018squeeze} recalibrates feature maps by capturing channel-wise dependencies, which has become a paradigm of channel attention. 
ECANet \cite{Wang2020ECANetEC} uses 1D convolution to capturing local cross-channel interaction while alleviating the negative impact of channel dimension reduction, and 
SRM \cite{lee2019srm} combines an attention mechanism with style pooling to extract style information from each channel and then recalibrates the weights of each channel through channel-independent style integration. 
The GCT \cite{ruan2021gaussian} employs a Gaussian function to directly map global contexts to attention activation, which significantly improves the generalization of the model. 

On the other hand, spatial attention can be interpreted as an adaptive spatial region selection mechanism. For example, NLNet \cite{wang2018non} utilizes non-local spatial structure, which can be viewed as a form of self-attention mechanism\cite{vaswani2017attention}, to capture long-range dependencies. A2-Net \cite{chen20182} leverages the same structure as NLNet capture long-range relation, but with different details in dimension and computation process.
Meanwhile, GcNet \cite{cao2019gcnet} simplifies the non-local block and incorporates an excitation block from SENet \cite{hu2018squeeze} to   enhance channel interaction. Lastly, SKNet \cite{li2019selective}  utilizes selective kernel unit to capture multiple scale feature information for channel representation.

Additionally, there are methods that combine both previous two ways. For instance, BAM \cite{park2018bam} and CBAM \cite{woo2018cbam} integrate both channel and spatial attention mechanisms and the previous method use an additive operator to fuse the two blocks, while the latter uses a  cascaded  operator for integration.

\subsubsection{Moment Statistics for Deep Learning}
Moment statistics is a potent tool for computer vision, including image moment \cite{flusser2000independence,flusser2006rotation}, 
statistical moment \cite{gonzalez2008digital1}, color moment \cite{smeulders2000content,yu2002color}, and moment texture approaches \cite{gonzalez2008digital2}, etc. 
In the image moment and color moment method, multiple moments are used to represent the image and color channels. Notably, the integration of moment estimation into deep learning has gained traction in recent years. Moment estimation method is applied for optimization algorithm \cite{Kingma2015AdamAM, Tong2022CalibratingTA, Zou2019ASC} and self-supervised learning \cite{li2021momentum}. Moreover, Zellinger et al. introduce a novel regularization technique, Central Moment Discrepancy (CMD), to facilitate domain adaptation from source to target domains \cite{zellinger2017central}. \looseness=-1

\begin{figure*}
	\centering
	\includegraphics[width= 13.5 cm]{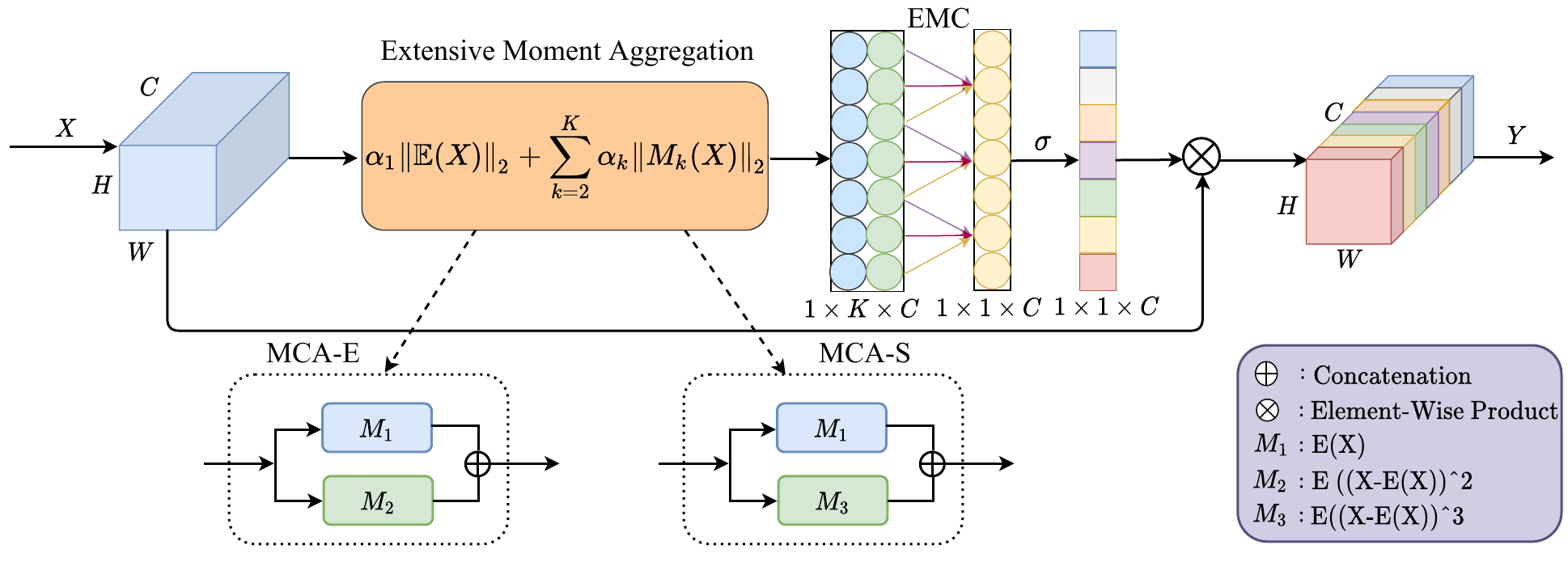}  %
	\caption{ Diagram of the moment channel attention (MCA) networks. We design a extensive moment aggregation (EMA) mechanism to capture global spatial feature, while the cross moment convolution (CMC) method facilitates cross-channel interactions between lower-order and other-order moments as well as different channels. The parameter $\alpha$ denotes as a learnable factor for each moment and $\sigma$ denotes Sigmoid function. $K$ indicates the order of moments, for MCA-E and MCA-S, $K$ is set to 2. 
	}
	\label{fig:fig2}
\end{figure*}

\section{The Proposed Approach}

In this section, we begin by introducing a novel moment aggregation mechanism based on probability distribution. Subsequently, we present our moment channel attention method.

\subsection{Extensive Moment Aggregation Mechanism}
  
We define the moment representation as an empirical estimate of the moment representation. Only the moments that correspond to the marginal distributions are computed. 

\begin{defi}[Extensive Moment Aggregation]

Let $X$ be bounded random samples with probability distribution $p$. The extensive moment aggregation $\text{EMA}_{K}$ is defined as an empirical estimate of the moment representation, by 
\begin{equation}
\text{EMA}_K(p)= \alpha_{1} \left \| {E(X)} \right \|_{2} + \sum_{k=2}^{K}  \alpha_{k} \left \| M_{k}(X) \right \|_{2}, 
\label{eqn:eq1}
\end{equation}
where $E(X)$ is the empirical expectation on the sample $X$ and $M_{k}(X) \!=\! E((X - E(X))^{k}$) encompasses all $k$-th order central moments of $X$. And $K$ serves as a parameter that restricts the quantity of moments computed, $\alpha_{k} \in (0, 1)$ act as weighted parameters. 
\end{defi}

The bound on the order of moment terms is limited by the parameter $K$. Proposition 1 demonstrates that the marginal distribution moment terms have an upper bound that strictly decrease with increasing  moment order.  
The theoretical analysis in later section shows that the marginal utility of higher-order terms fade as the order increases. Specifically, within this framework, we can always find  a best trade-off between distribution approximation and computational efficiency. 

\begin{prop} [Upper Bound]
Let $X$ be bounded random vector with probability distribution $p$ on compact interval $[a, b]^{N}$. Then, for all positive integers $k$, 
\begin{equation}
\alpha_{k} \left \| M_{k}(X) \right \|_{2} \le  \sqrt{N} \left (\frac{1}{k+1}(\frac{k}{k+1})^{k}+ \frac{1}{2^{1+k}}  \right ). 
\label{eqn:eq2}
\end{equation}
where $M_{k}(X) = E((x - E(X))^{k})$ is the vector of all $k$-th order moments of the marginal distributions of $p$. And $\alpha_{k} \in (0, 1)$ act as weighted parameters. 
\end{prop}

\begin{proof}
	Let $X([a,b]) $ denote the set of all random variables with values in the interval $[a,b]$. Then, it follows that
	
	\begin{equation}
	\begin{aligned}   
	\alpha_{k}  \left \| M_{k}(X) \right \|_{2} 
	& \le \left \| \mathbb{E} \left ( \left (\left | \frac{X-\mathbb{E}(X)}{b-a}  \right | \right )^{k}\right ) \right \|_{2}    \\
	& \le \sqrt{N}\sup_{X\in \mathcal{X}([a,b])} \mathbb{E} \left (\left | \frac{X-\mathbb{E}(X)}{b-a}  \right |^{k} \right ).
	\end{aligned}
	\end{equation}
	
	Here, the latter term refers to the absolute moment of order $k$, for which the smallest upper bound is known (Egozcute et al. \cite{egozcue2012smallest}):
	\begin{equation}
	\alpha_{k} \left \| M_{k}(X) \right \|_{2} \le 
	\sqrt{N} \sup_{X\in \mathcal{X}([0,1])} x(1-x)^{k} + (1-x)x^{k}.
	\end{equation}
	Then, we can get a more explicit upper bound:
	
	\begin{equation}
	\alpha_{k} \left \| M_{k}(X) \right \|_{2} \le \sqrt{N} \left (\frac{1}{k+1}(\frac{k}{k+1})^{k} + \frac{1}{2^{1+k}}  \right ). 
	\end{equation}
	
\end{proof}

The proof of theorem and proposition can also refer to Zellinger et al. \cite{zellinger2017central,zellinger2019robust} for detail.

The equation presented in Eq.~\eqref{eqn:eq1} offers a comprehensive formalization of moment information aggregation, and Eq.~\eqref{eqn:eq2} demonstrates that  a constrained set of moments can effectively  represent the distribution. By controlling the parameter $K$, we can readily balance between distribution approximation accuracy and computational efficiency. 

We apply the extensive moment aggregation for the global feature extraction of channel attention mechanism and propose a novel attention method.

\subsection{Moment Channel Attention Block}
In this section, we present the architecture of MCA in detail. The MCA method comprises three modules: (a) moment aggregation; (b) moment fusion; (c) channel recalibration. The overall framework of MCA is illustrated in Fig.~\ref{fig:fig2}.

\subsubsection*{Moment Aggregation}
  
From extensive moment aggregation $\textit{EMA}_{K}$,
it is evident that moments consist of first-order moment $M_{1}$, other-order moments along with their combination. So for $M_{1}$, $M_{1}\!=\!E(X)$; and for other moments, $M_{k}(X) = E(( X - E(X))^{k}) (k \ge 2)$.

As showed in $\textit{EMA}_{K}$, The moment aggregation module can obtain single moment: the first-order moment $M_{1}$ or the second-order moment $M_{2}$ or the third-order moment $M_{3}$ for each channel. so the $M_{mono}$ is represented as follows:
\begin{equation}
M_{mono} = \left[M_{1} | M_{2} | M_{3}\right].
\label{eqn:eq3}
\end{equation}

$M_{mono}$ is a refined representation  for this channel and $ M_{mono} \in \mathbb{R}^{1 \times 1 \times C}$. $M_{mono}$ will out to the next layer as the basis for weight recalibration.

In the same way, the ${M}_{dual}$ module combines two of $M_{1}$, $M_{2}$ and $M_{3}$ as the aggregation of global feature and ${M}_{dual} \in \mathbb{R}^{1 \times 2 \times C}$.  The ${M}_{dual}$ is represented as follows:
\begin{equation}
{M}_{dual} =  \left[M_{1} \wedge M_{2}]  | [M_{1}\wedge M_{3}] | [M_{2} \wedge M_{3}\right].
\label{eqn:eq4}
\end{equation}

In addition, the ${M}_{triple}$ obtains $M_{1}$,  $M_{2}$ and $M_{3}$ for each channel as the representation of the channel and  ${M}_{triple} \in \mathbb{R}^{1 \times 3 \times C}$. The specific calculation of ${M}_{triple}$ is as follows:
\begin{equation}
{M}_{triple} =  \left[M_{1} \wedge M_{2} \wedge M_{3}\right].
\label{eqn:eq5}
\end{equation}

To summarize, the steps of ${M}_{mono}$, ${M}_{dual}$, and ${M}_{triple}$ represent the detail implementation of extensive moment aggregation. Both ${M}_{dual}$ and ${M}_{triple}$ result in an increase in dimension, which is a matter we address in the subsequent section.

\subsubsection*{Moment Fusion}

Next, we delve into moment fusion approach aimed at resolving the dimension issue associated with extensive moment aggregation $\textit{EMA}_{K}$  and this method effectively captures both the low-order and other-order moment information across the channel.

Note that the dimension of ${M}_{mono}$, ${M}_{dual}$, and ${M}_{triple}$ are different. How to fuse the moment vector is a challenge. In the SENet method and the ECANet method, the dimension of the vector is also $1\times 1\times C$, and they are processed by a fully connected \cite{hu2018squeeze} layer and a Conv1D \cite{Wang2020ECANetEC} layer, respectively. In the SRM method, the dimension of input is $1 \times 2 \times C$, they use the channel-wise fully connected (CFC) \cite{lee2019srm} layer for fusion. During this study, we employ cross moment convolution (CMC) method, which utilizes channel-wise 1D convolution layer for fusion multiple moments. The input vectors correspond to the output of either ${M}_{mono}$, ${M}_{dual}$, or ${M}_{triple}$., and the output $F \in \mathbb{R}^{1 \times 1 \times C}$. the moment fusion is formulated as:
\begin{equation}
 F  =  \textrm{CMC}(M).
\label{eqn:eq6}
\end{equation}

The CMC method fuses information cross the channel through the convolutional kernel, and can fuse different orders moment information in the channel.

\subsubsection*{Channel Recalibration}

Finally, through the above moment fusion method, we obtain the extensive moment feature aggregation. The entire algorithm can be represented in a unified form. The entire MCA block of is expressed as follows:
\begin{equation}
Y =  X \cdot \sigma( F ).
\label{eqn:eq7}
\end{equation}

Among them, $X \in \mathbb{R}^{C \times H \times W}$ be an activation feature in a convolutional network, where $C$, $H$, and $W$ denote the number of channels, the spatial height and width, respectively. And $M$ can be ${M}_{mono}$, ${M}_{dual}$ or ${M}_{triple}$. The dimension of CMC is adjusted according to the dimension of Moment.

Theoretical analysis of extensive moment aggregation indicates that the performance improves as the moment order increase. This observation is further validated by the results of  ablation studies.  
Specifically, $M_2$ and $M_3$ demonstrates better than $M_1$, and ${M}_{dual}$ outperforms ${M}_{mono}$. 
While ${M}_{triple}$ exhibits superiority over ${M}_{dual}$. 
It is noteworthy that ${M}_{triple}$ slightly outperforms performance to $[M_{1} \wedge M_{3}]$. 
It can be seen that the extensive moment aggregation is an effective way of extracting representative information.

In this paper, we conduct the method using an extensive moment aggregation manner, fundamental term and additional terms. 
Additionally, computational complexity and performance are also considerations. Taking these factors into account, we have chosen two methods, namely Moment-E and Moment-S, for further investigation.
In the Moment-E method, we adopt $M = [M_{1} \wedge M_{2}]$, while in the Moment-S method, we utilize $M = [M_{1} \wedge M_{3}]$. Subsequent experiments and analyses are conducted based on the aforementioned two algorithms.

\subsection*{Parameter and Computational Complexity}

This section analyzes the parameter and computational complexity of the MCA algorithm. 
The number of parameters for MCA(MCA-E, MCA-S) is
$ \sum_{s=1}^{S}N_{s}\cdot C_{s} *2$, where $S$ denotes the numbers of stages,  $N_{s}$ is  the number of repeated blocks in s-th stage, and $C_{s}$ is the dimension of the output channels for s-th stage. 
While the extra parameters of SENet is $\frac{2}{r} \sum_{s=1}^{S}N_{s}\cdot C_{s}^{2}$,  where r is its reduction ratio. For instance, given ResNet-50 as a baseline for image classification, MCA-E and MCA-S require only 6.06K additional parameters whereas SENet requires 2.53M. MCA also introduces negligible extra computations to the original architecture.  
A single forward pass of a $224 \times 224 $ pixel image for MCA-S requires additional 0.011 GFLOPs to ResNet-50 which requires 4.122 GFLOPs. 
By adding only 0.27\% relative computational burden, MCA-E increases the Top-1 accuracy from 74.97\% to 76.61\%, which indicates that MCA offers a good trade-off between accuracy and efficiency. 

\section{Experiments}

In this section, we evaluate the performance of the MCA method on the COCO and ImageNet datasets, along with other channel attention methods, such as SENet \cite{hu2018squeeze}, SRM \cite{lee2019srm}, ECANet \cite{Wang2020ECANetEC}, and GCT method \cite{yang2020gated}. 
To ensure a fair comparison, we use the same network architecture, data augmentation strategy, and optimization parameters for all methods without any additional bells and whistles.

\subsection{Implementation Details}

We evaluate the effectiveness of our approach in object detection and instance segmentation tasks using the COCO \cite{lin2014microsoft} dataset. 
The training of all models is performed on 8 GPUs, with each mini-batch consisting of eight images. As part of the preprocessing procedure, the shorter edge of the input image is resized to 800 while the longer edge is constrained to a maximum size of 1333 and RandomFlip is used for  data augmentation. 
In terms of optimization, stochastic gradient descent (SGD) was chosen as the optimizer,
with a weight decay of 1e-4, and a momentum of 0.9. All models are trained for a total of 12 epochs. 
The learning rate is set to 0.02 initially and drops to 0.002 and 0.0002 at the 8th and 11th epochs, respectively.
For the object detection and instance segmentation tasks, we utilize the mean average precision (AP), $AP_{0.5}$, $AP_{0.75}$, $AP_{S}$, $AP_{M}$, and $AP_{L}$ as the evaluation metrics. Additionally, we also test the storage capacity by Params, as well as the computing efficiency in GFLOPs.

Next, we further evaluate the MCA method for image classification tasks on the ImageNet dataset \cite{russakovsky2015imagenet}.
All models are trained with a mini-batch size of 256 using 8 GPUs (32 images per GPU).
The size of the image is 224$\times$224, and we use RandomFlip in the horizontal direction for data augmentation. 
In terms of optimization, SGD was selected as the optimizer,
with a weight decay of 1e-4, and a momentum of 0.9. The learning rate is initial to 0.1. 
The model is trained for 100 epochs and the learning rate is decreased by a factor of 0.1 at epochs 30, 60, 90. We insert the attention module after the third convolutional layer of each residual module.
For the image classification experiment, we use Top-1 and Top-5 accuracy as the test metrics. Additionally, we evaluate the storage capacity with Params and computing efficiency in terms of GFLOPs.

\subsection{Ablation Study}

First, we perform ablation analysis to investigate the effects of the moments selection and coverage of moment cross-channel interaction. The Faster RCNN \cite{ren2015faster} is utilized as the base detector for experimenting on the COCO dataset. The ResNet-50 is employed as the backbone model. 

\subsubsection*{The Selection of Moments}
The aim of our proposed method is to conduct aggregation of extensive moment feature, so we conduct ablation experiments to explore the impact of moments on the performance, specifically on their selection.
Moments include first-order moment $M_{1}$, second-order moment $M_{2}$, third-order moment $M_{3}$, and fusion versions include $M_{1}+M_{2}$, $M_{1}+M_{3}$, $M_{2}+M_{3}$,
and $M_{1}+M_{2}+M_{3}$. Table~\ref{tbl:table2} shows that ${M}_{triple}$ ($M_{1}+M_{2}+M_{3}$) has the best performance in the object detection tasks, followed by the MCA-S ($M_{1}+M_{2}$) method. 
Several key observations  can be derived from Table~\ref{tbl:table2}. Firstly, higher-order moments exhibit superior performance compared to lower-order moments. Notably, $M_3$ outperforms both $M_2$ and $M_1$. Secondly, the combination of multiple moments yields superior results compared to using individual moment. For instance, the combination of $M_1$+$M_2$ performs better than employing either $M_1$ or $M_2$ alone. Additionally, the combination of $M_1$+$M_3$ surpasses the performance achieved by using $M_1$ or $M_3$ individually. Thirdly, ${M}_{triple}$ demonstrates a clear advantage over ${M}_{dual}$ and ${M}_{mono}$ in terms of performance. Fourthly,  complex combinations of moments ${M}_{triple}$ show  limited improvement. 
Although extensive moment aggregation has an upper bound $k$. Considering the computational complexity, we do not investigate further to the fourth moment Kurtosis \cite{bai2005tests}. 
\begin{table}[t]
\small
\centering
\setlength{\tabcolsep}{4pt}
\begin{tabular}{ c|cccccccc }

\hline

 \multicolumn{3}{c|}{Methods}  & \multirow{2}{*} {AP} & \multirow{2}{*} {$AP_{0.5}$} & \multirow{2}{*} {$AP_{0.75}$} & \multirow{2}{*} {$AP_{S}$} & \multirow{2}{*} {$AP_{M}$}  & \multirow{2}{*} {$AP_{L}$} \\ \cline{1-3}

 \multicolumn{1}{c|}{$M_{1}$} & \multicolumn{1}{|c|}{$M_{2}$} &\multicolumn{1}{|c|}{$M_{3}$}  &   &  & &  & &  \\ 
 \hline
 \multicolumn{1}{c|}{} & \multicolumn{1}{|c|}{} &\multicolumn{1}{|c|}{} & 34.9 & 56.6 & 37.1 & 20 & 38.5& 45.6 \\
 \multicolumn{1}{c|}{{\Checkmark}} & \multicolumn{1}{|c|}{} & \multicolumn{1}{|c|}{}  & 36.2 & 58.4 & 38.6 &  21.1 & 39.1 & 45.9  \\
 \multicolumn{1}{c|}{} &\multicolumn{1}{|c|}{{\Checkmark}}  & \multicolumn{1}{|c|}{}   & 37.6 & 59.4 & 40.6 & 21.9 & 41.0 & 48.3 \\
 \multicolumn{1}{c|}{} & \multicolumn{1}{|c|}{}  & \multicolumn{1}{|c|}{{\Checkmark}}  & 38.1& 60.0 & 41.4 & 23.1 & 42.1 & 48.8\\ 
 \multicolumn{1}{c|}{ {\Checkmark}} & \multicolumn{1}{|c|}{{\Checkmark}}  &  \multicolumn{1}{|c|}{}  & 38.0 & 60.0 & 40.7 & 22.4 & 42.1 & 48.2  \\ 
 \multicolumn{1}{c|}{ {\Checkmark}} & \multicolumn{1}{|c|}{}  & \multicolumn{1}{|c|}{{\Checkmark}}  &38.3 & \textbf{60.5} & 41.4 & 22.6 & \textbf{42.4} & 49.4  \\ 
 \multicolumn{1}{c|}{} & \multicolumn{1}{|c|}{{\Checkmark}}  & \multicolumn{1}{|c|}{{\Checkmark}}  & 38.2 & 60.2 & 41.5 & 22.8 & 42.1 & \textbf{49.5} \\
 \multicolumn{1}{c|}{{\Checkmark}} & \multicolumn{1}{|c|}{{\Checkmark}}  & \multicolumn{1}{|c|}{{\Checkmark}}  & \textbf{38.4} & 60.7 & \textbf{41.6} & \textbf{23.2} & 42.2 & 49.3 \\
\hline
\end{tabular}
\caption{Ablation study of different moments on COCO val2017 set with object detection task and the baseline model is Resnet-50.}
\label{tbl:table2}
\end{table}

\subsubsection*{Coverage of Moment Cross-Channel Interaction}
\begin{figure}[ht]
\centering
\includegraphics[width=8.0 cm]{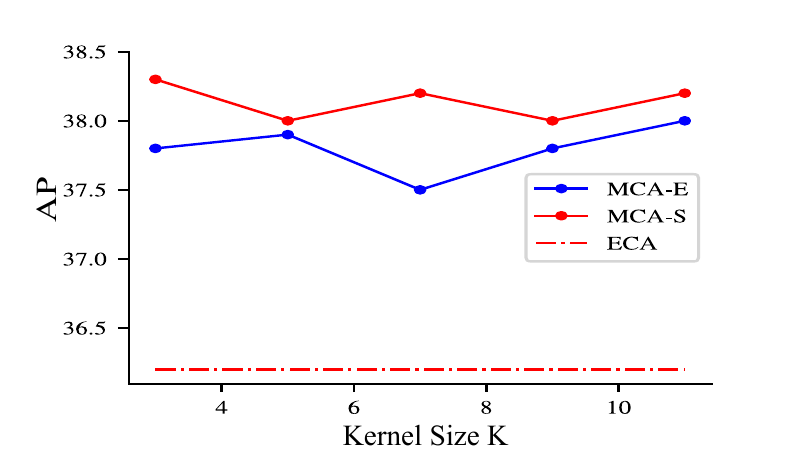} 

\caption{ Results of MCA-E and MCA-S with various kernel size $k$ for 3 to 11 using ResNet-50 as backbone model. Here we choose the ECA method as the baseline. }
\label{fig:fig3}
\end{figure}

The another aim of our MCA method is capturing moment cross-channel interaction appropriately, so the coverage of interaction (i.e., kernel size $k$ of CMC method) needs to be determined. 
In our experiments, we set the kernel size $k$ for MCA-E and MCA-S to range from 3 to 11 for simplicity.
The results are depicted in Fig.~\ref{fig:fig3}, from which we have three observations.
(i) MCA block with different $k$  outperforms ECANet block, demonstrating the effectiveness of utilizing moments in channel attention.
(ii) The performance of the MCA-E method has a zigzagging upward trend with slight fluctuations; the performance of the MCA-S method has a zigzagging downward trend with slight fluctuations.
(iii) For MCA-E and MCA-S, the setting with 11 and 3   achieves the best performance, respectively.  In this way, we use these settings in our block and other experiments.

\subsection{Object Detection on COCO Dataset}

We also conduct object detection experiments on the COCO dataset and compared the performance of our method with SENet, SRM, ECANet, and GCT methods. The experimental results are shown in Table~\ref{tbl:table4}. For object detection, we use three types of detectors, namely Faster RCNN \cite{ren2015faster},
Mask RCNN \cite{he2017mask}, and RetinaNet \cite{lin2017focal}. The ResNet-50 and ResNet-101 models are used as the backbone models. 

Overall, the performance of MCA-E and MCA-S is superior to that of other attention blocks on the object detection task.
In the Faster RCNN experiment, with ResNet50 as the backbone model, compared with the baseline, SENet, SRM, ECANet, and GCT methods, the MCA method exhibits AP improvements of 3.2\%, 2.9\%, 1.4\%, 2.1\%, and 0.4\%, correspondingly; Subsequently, using employing ResNet-101 as the backbone model, the MCA method achieves superior AP compared to the baseline, SENet, SRM, ECANet, and GCT method, with improvement of 1.3\%, 1.2\%, 1.0\%, 0.4\%, and 0.2\%, respectively.
In the Mask RCNN  experiment, utilizing ResNet-50 as the backbone model, compared with the baseline, SENet, SRM, ECANet, and GCT blocks, the MCA method improves the AP by 1.5\%, 1.2\%, 1.0\%, 0.9\%, and 0.5\%, respectively; Similarly, with ResNet-101 as the backbone, the AP of MCA methods outperforms the baseline, SENet, SRM, ECANet, and GCT by 1.3\%, 0.9\%, 1.4\%, 0.8\%, and 0.2\%, respectively.
In the RetinaNet experiment, using ResNet-50 is the backbone model, compared with the baseline, SENet, SRM, ECANet, and GCT methods, the MCA method yields the AP by 1.1\%, 1.4\%, 0.3\%, 1.0\%, and 0.3\%, respectively; with ResNet-101 as the backbone, the MCA method surpasses the performance of the baseline, SENet, SRM, ECANet, and GCT methods by 1.2\%, 1.4\%, 2.0\%, 0.8\%, and 0.5\%, respectively. 

\begin{table}
	\footnotesize
	\centering 
	\setlength{\tabcolsep}{4pt}
		\begin{tabular}  { l|cccccc }
			
			\hline
			Methods & AP & $AP_{0.5}$ & $AP_{0.75}$ & $AP_{S}$ & $AP_{M}$ & $AP_{L}$ \\ \hline
			
			ResNet50 & 34.0 & 55.2 & 36.0 & 16.3 & 36.9 & 49.6 \\
			+SE  & 34.2 & 56.0 & 36.0 & 15.9 & 37.2 & 49.8  \\
			+SRM  & 34.5 & 55.4 & 36.8 & 16.0 & 36.9 & 50.7 \\
			+ECA & 34.5 & 56.4 & 36.4 & 16.1 & 37.7 & 50.3 \\
			+GCT & 34.8 & 56.8 & 37.1 & 16.2 & 37.8 & 50.5\\ 
			+MCA-E & 35.3 & \textbf{57.6} & 37.3 & \textbf{16.8} & 38.4 & \textbf{51.4}  \\ 
			+MCA-S & \textbf{35.4} & 57.4 & \textbf{37.7} & 16.7 & \textbf{38.7} & 50.7  \\ 
			\hline
			ResNet101  & 35.7 & 57.2 & 38.0  & 16.6 & 38.7 & 52.2  \\
			+SE       & 36.0 & 58.1 & 38.2 & 17.7 & 39.2 & 52.6 \\
			+SRM  & 35.8 & 57.2 & 38.0 & 16.1 & 39.0 & 52.4 \\
			+ECA   & 36.2 & 58.2 & 38.5 & 17.6 & 39.4 & 53.3\\ 
			+GCT     & 36.7 & 59.0 & \textbf{38.9} & 17.4 & 40.1 & \textbf{53.6} \\ 
			+MCA-E  &36.6 & 59.0 &\textbf{38.9}  & 17.3 & 39.9 & 53.5\\ 
			+MCA-S  & \textbf{36.8} & \textbf{59.5} & 38.8  & \textbf{18.0} & \textbf{40.5} & 53.3\\ 
			\hline
		\end{tabular}
	\caption{Instance Segmentation of the state-of-the-art channel attention blocks on COCO val2017 set. }
	\label{tbl:table3}
\end{table}

\begin{table*}[t]
	\small
	
	\centering
	\setlength{\tabcolsep}{4pt}
		\begin{tabular}	{ c|l|lc|cccccc }
			\hline
			Detector & Methods &  Params & GFLOPs & AP & $AP_{0.5}$ & $AP_{0.75}$ & $AP_{S}$ & $AP_{M}$ & $AP_{L}$ \\ \hline
			\multirow{14}{*}{\centering Faster \ RCNN} & ResNet-50 & 41.53M & 207.07 & 34.9 & 56.6 & 37.1 & 20 & 38.5& 45.6 \\
			\multirow{8}{*}{}& +SE\quad (CVPR,2018)  & +2.49M & 207.18  & 35.4 & 57.4 & 37.7&  20.8& 39.1 & 45.9  \\
			\multirow{8}{*}{}& +SRM (ICCV,2019)  & +0.06M & 207.07  & 36.9& 57.5 & 39.8 & 21.1 & 40.4 & 47.0  \\ 
			\multirow{8}{*}{}& +ECA (CVPR,2020) & +0.05k & 207.18 & 36.2& 58.4 & 38.6 & 21.1 & 41.8 & 49.0 \\
			\multirow{8}{*}{}& +GCT (CVPR,2021)  & +0.01k & 207.18  & 37.9& 59.6 & 41.1 & 22.3 & 40.7 & 47.9  \\ 
			\multirow{8}{*}{}& +MCA-E (Ours) & +0.03M & 207.07  & 38.0& 60.0 & 40.7 & 22.4 & 42.1 & \textbf{49.6}  \\ 
			\multirow{8}{*}{}& +MCA-S (Ours) & +0.03M & 207.07  & \textbf{38.3} & \textbf{60.5} & \textbf{41.4} & \textbf{22.6} & \textbf{42.4} & 49.4  \\ 
			
			\cline{2-10}
			\multirow{8}{*}{} &ResNet-101 & 60.52M & 283.14   & 39.0& 60.2 & 42.1  & 22.2 & 43.0 & 51.0  \\
			\multirow{8}{*}{} & +SE\quad (CVPR,2018)  & +4.72M & 283.33      & 39.1 & 60.5 & 42.3 & 22.1 & 43.1 & 51.3 \\
			\multirow{8}{*}{} & +SRM (ICCV,2019) & +0.13M & 283.14     & 39.0 & 59.6 & 42.3 & 22.0 & 42.9 & 50.6  \\ 
			\multirow{8}{*}{} & +ECA (CVPR,2020)   & +0.12k & 283.32     & 39.9 & 61.3 & 43.4 & 22.7 & 44.2 & 52.1\\ 
			\multirow{8}{*}{} & +GCT (CVPR,2021)  & +0.03k & 283.32     & 40.1 & 61.7 & 43.5 & 22.9 & 42.7 & 50.5  \\ 
			\multirow{8}{*}{} & +MCA-E (Ours) & +0.06M & 283.14 &\textbf{40.3} & 61.9 & \textbf{43.7} & 23.5 & 44.4 & \textbf{52.5}  \\ 
			\multirow{8}{*}{} & +MCA-S (Ours) & +0.06M & 283.14   & 40.2 & \textbf{62.3} &43.2  & \textbf{23.6} & \textbf{44.6} & 52.2  \\ 
			\hline
			\multirow{14}{*}{\centering Mask \ RCNN} & ResNet-50 & 44.17M & 260.14 & 37.5 & 58.7 & 40.3 & 22.2 & 40.9 & 48.8 \\
			\multirow{8}{*}{}& +SE\quad (CVPR,2018)  & +2.49M & 260.25  & 37.8 & 59.2 & 40.7 &  22.2 & 41.5 & 48.7  \\
			\multirow{8}{*}{}& +SRM (ICCV,2019)   & +0.05M & 260.14 & 38.0 & 58.7 & 41.6 & 21.9 & 41.3 & 49.8 \\
			\multirow{8}{*}{}& +ECA (CVPR,2020)  & +0.05k & 260.25  & 38.1 & 59.8 & 41.0 & 22.1 & 42.1 & 49.4 \\
			\multirow{8}{*}{}& +GCT (CVPR,2021) & +0.01k & 260.25  & 38.5 & 60.1 & 41.8 & 22.4 & 42.4 & 49.7 \\
			\multirow{8}{*}{}& +MCA-E (Ours) &+0.03M & 260.14 & \textbf{39.0} & \textbf{61.0} & 42.3 & 22.9 & 42.7 & \textbf{50.5}  \\ 
			\multirow{8}{*}{}& +MCA-S (Ours) &+0.03M & 260.14  & \textbf{39.0}& 60.9 & \textbf{42.7} & \textbf{23.3} & \textbf{42.9} & \textbf{50.5}  \\
			\cline{2-10}
			\multirow{8}{*}{} &ResNet-101 & 63.16M & 336.21  & 39.7 & 60.6 & 43.3 & 22.9 & 43.8 & 52.2  \\
			\multirow{8}{*}{} & +SE\quad (CVPR,2018)  & +4.72M & 336.40        & 40.1 & 61.4 & 43.4 & 23.6 & 44.2 & 52.4 \\
			\multirow{8}{*}{} & +SRM (ICCV,2019)  & +0.12M & 336.21     & 39.6 & 60.6 & 43.0 & 22.4 & 43.8 & 52.4 \\ 
			\multirow{8}{*}{} & +ECA (CVPR,2020)  & +0.17k & 336.39          & 40.2 & 61.6 & 43.9 & 23.6 & 44.5 & 52.7\\ 
			\multirow{8}{*}{} & +GCT (CVPR,2021)  & +0.03k & 336.39          & 40.8 & 62.4 & 44.4 & 24.1 & 45.0 & 53.2\\ 
			\multirow{8}{*}{} & +MCA-E (Ours) & +0.07M & 336.21    & 40.8 & 62.4 & 44.3 & 23.7 & 45.0 & \textbf{52.9}  \\ 
			\multirow{8}{*}{} & +MCA-S (Ours) & +0.07M & 336.21    & \textbf{41.0} & \textbf{62.8} & \textbf{44.8} & \textbf{24.9} & \textbf{45.1} & \textbf{52.9}  \\
			\hline
			\multirow{14}{*}{ \centering RetinaNet} & ResNet-50 & 37.74M & 239.32 & 35.0 & 54.2 & 37.1 & 20.4 & 38.6 & 45.4 \\
			\multirow{8}{*}{}& +SE\quad (CVPR,2018)  & +2.49M & 239.43  & 34.7 & 57.4 & 37.7&  20.8 & 39.1 & 45.2  \\
			\multirow{8}{*}{}& +SRM (ICCV,2019)   & +0.06M & 239.32 & 35.8 & 54.8  & \textbf{38.3} & 20.3 & 39.6 & 46.2 \\
			\multirow{8}{*}{}& +ECA (CVPR,2020)  &  +0.09k & 239.43 & 35.1 & 54.8 & 37.1 & 20.5 & 38.7 & 45.8 \\
			\multirow{8}{*}{}& +GCT (CVPR,2021) & +0.01k & 239.43 & 35.8 & 55.6 & 37.9 & 20.5 & 39.6 & 46.6 \\
			\multirow{8}{*}{}& +MCA-E (Ours) &+0.03M & 239.32  & 35.4 & 55.0 & 37.6 & 20.3 & 39.1 & 46.2  \\ 
			\multirow{8}{*}{}& +MCA-S (Ours) &+0.03M & 239.32  & \textbf{36.1}& \textbf{55.9} & 38.0 & \textbf{21.1} & \textbf{39.9} & \textbf{47.2}  \\ 
			\cline{2-10}
			\multirow{8}{*}{} &ResNet-101 & 56.74M & 315.39   & 37.6 & 57.0 & 40.4  & 21.3 & 42.1 & 48.9  \\
			\multirow{8}{*}{} & +SE\quad (CVPR,2018) & +4.71M & 315.58      & 37.4 & 56.7 & 40.2 & 20.9 & 41.8 & 48.5 \\
			\multirow{8}{*}{} & +SRM (ICCV,2019)   &  +0.12M & 315.39          & 36.8 & 55.9 & 39.2 & 20.7 &  41.1 & 47.8 \\ 
			\multirow{8}{*}{} & +ECA (CVPR,2020) &  +0.17k & 315.57      & 38.0 & 57.7 & 40.3 & 21.6 & 42.2 & 49.1\\ 
			\multirow{8}{*}{} & +GCT (CVPR,2021  &  +0.03k & 315.57        & 38.3 & 58.0 & 41.0 & 21.9 & 42.4 & 49.6\\ 
			\multirow{8}{*}{} & +MCA-E (Ours) & +0.06M & 315.39  & 38.0 & 57.9 & 40.4  & \textbf{22.2} & 42.2 & 49.5  \\ 
			\multirow{8}{*}{} & +MCA-S (Ours) & +0.06M & 315.39   & \textbf{38.8} & \textbf{58.9} & \textbf{41.4}  & 21.7 & \textbf{43.0} & \textbf{51.3}  \\ 
			\hline
			
		\end{tabular}
	\caption{ Comparisons between different methods on the COCO val2017 set with object detection task.}
	\label{tbl:table4}
\end{table*}

\begin{table}[t]
\small
\centering
\begin{tabular}{ l|ll|ll }

  \hline
  Methods & Params & GFLOPs & Top-1 & Top-5 \\ \hline
  ResNet-18 & 11.69M & 1.822 & 69.76 & 89.08 \\
  +SE  & +89.08K & 1.823 & 70.59 & 89.78 \\ 
  +SRM    & +0.77K  & 1.823   & 69.89 & 89.45\\ 
  +ECA    & +0.04K & 1.823 & 70.85 & 89.75 \\
  +GCT   & +0.01K  & 1.823   & 71.21 & 90.04 \\  
  +MCA-E  & +0.77K  & 1.823   & \textbf{71.23} & \textbf{90.10} \\ 
  +MCA-S  & +0.78K  & 1.823   & 70.93 & 89.87 \\ 
  \hline
  ResNet-50 & 25.56M & 4.122 & 74.97 & 92.23 \\
  +SE   & +2.53M & 4.130 & 75.90 & 92.75 \\
  +SRM    & +6.04K  & 4.122   & 76.44 & 93.02\\  
  +ECA    & +0.09K & 4.127 & 75.40 & 92.66 \\
  +GCT   & +0.02K & 4.127 & 74.02 & 91.75 \\ 
  +MCA-E  & +6.06K & 4.122 & 76.36  & 93.14 \\ 
  +MCA-S & +6.06K & 4.133 & \textbf{76.61} & \textbf{93.21} \\ 
  \hline
  ResNet-101 & 44.55M & 7.849 & 76.51 & 93.10 \\
  +SE    & +4.78M & 7.863 & 76.72 & 93.31 \\
  +SRM    & +130.0K  & 7.849   & 77.96 & 93.92 \\   
  +ECA    & +0.16K & 7.858 & 76.56 & 92.96 \\ 
  +GCT    & +0.03K & 7.858 & 75.66 & 92.61 \\ 
  +MCA-E & +130.4K & 7.849 & 78.18 & \textbf{93.97} \\ 
  +MCA-S & +130.4K & 7.858 & \textbf{78.21} & 93.88 \\ 
  \hline
  ShufflenetV2 & 2.28M & 0.151 & 65.71 & 86.74 \\
  +SE  & +16.8K & 0.152 & 66.95 & 87.68 \\ 
  +SRM    & +16.7K  & 0.151   & 67.32  & 87.41 \\  
  +ECA    & +0.08K & 0.151 & 66.56 & 87.03 \\ 
  +GCT   & +0.02K & 0.151 & 62.24 & 84.20 \\ 
  +MCA-E & +16.7K & 0.151 & \textbf{67.93} & 87.87 \\ 
  +MCA-S & +16.9K & 0.151 & 67.91 & \textbf{88.00} \\ 
  \hline

\end{tabular}
\caption{Image classification results of the state-of-the-art channel attention blocks on ImageNet dataset.}

\label{tbl:table5}
\end{table}

\subsection{Instance Segmentation on COCO Dataset}
Next, we conduct instance segmentation experiments utilizing the COCO dataset, comparing proposed MCA method with other blocks including SENet, SRM, ECANet, and GCT method. 
we utilize  Mask RCNN as the detector and select ResNet-50 and ResNet-101 models as the backbone models.
Our proposed method outperforms the other channel attention method in terms of instance segmentation as the results shown in Table~\ref{tbl:table3}. 
(i) With the ResNet-50 as the backbone model, the MCA method enhances the performance of the baseline, SENet, SRM, ECA, and GCT methods by 1.4\%, 1.2\%, 1.2\%, 0.9\%, and 0.6\%, respectively. (ii) Using ResNet-101 as the backbone model, the MCA method outperforms the baseline, SENet, SRM, ECA, and GCT methods by 1.1\%, 0.8\%, 1.0\%, 0.6\%, and 0.1\%, respectively. These outcomes show the generalization and effectiveness of our MCA method.

\subsection{Image Classification on ImageNet Dataset}

Finally, we conduct image classification experiments on the ImageNet dataset. And we choose 4 models as the backbone, including: ResNet \cite{he2016deep} and ShuffleNetV2 \cite{ma2018shufflenet} model. The experimental results of image classification are shown in Table~\ref{tbl:table5}.

\subsubsection{ResNet} We first evaluate our MCA module on popular ResNet model. All attention modules are placed after the last BatchNorm layer inside each bottleneck of ResNet. MCA-E performs better than SENet, SRM, ECA, and GCT across different depths and backbones with similar parameters and slightly more computation. The results show that our MCA method can be used to improve the performance of residual network.
It is worth noting that the SRM method achieves the comparable performance with MCA-E on Resnet-50 model but have a poor performance on Reset-18 model, while GCT method has comparable performance with MCA-E on Resnet-18 model. In general, MCA method has better generalization performance and brings improvement to different types of models.

\subsubsection{ShufflenetV2} We also verify the effectiveness of our ECA module on lightweight CNN architectures. We utilize ShufflenetV2 \cite{ma2018shufflenet} as backbone model, comparing our MCA module with baseline, SENet, SRM, ECA, and GCT module. All attention modules are placed before channel shuffle. The results in Table~\ref{tbl:table5} show that MCA-E performance better than previous best SRM block by 0.61\% in Top-1 accuracy. Compared to vanilla ShufflenetV2, our MCA-E improves by a 2.21\% in Top-1 accuracy. These results indicate that MCA can be successfully applied to lightweight model.
  
\section{Conclusion}

In this paper, we analyze the excitation module of channel attention mechanism from a statistical perspective and utilize moment statistics to aggregate extensive moment feature, and propose the Moment Channel Attention (MCA) method that achieves higher performance and comparable complexity. Furthermore, we introduce the Cross Moment Convolution method to fuse low-order and other-order information both inside and cross the channel. Additionally, through a series of ablation experiments, we demonstrate the effectiveness of our MCA method. Finally, we evaluate the performance of our method in image classification, object detection, and instance segmentation tasks on large-scale datasets such as the COCO and ImageNet. The results of our comparative analysis show that our method outperforms other state-of-the-art channel attention methods. 
In addition, there are several aspects that need further improvement and investigation: 
(1) The  ${M}_{triple}$ method, as well as single moment $M_{2}$ and $M_{3}$ have not been thoroughly explored yet, leaving potential for further improvement in performance.
(2) The fourth moment Kurtosis \cite{bai2005tests} is widely used in the economics. Incorporate Kurtosis as another moment representation could be a valuable avenue for future investigation.
In the future work, we will further investigate incorporation of MCA with spatial attention module.
 
\appendix
\section{Appendix}

\subsection{Further Experiment Results}

\subsubsection* {Moment Fusion Method}

For the moment fusion module, only the SRM \cite{lee2019srm} method proposed the CFC method before, and the CFC method adopted the channel-wise fully connected layer for moment information fusion without considering the local channel interaction. CMC method is proposed in this paper, which adopts channel-wise Conv1D layer for information fusion, considering both moment information fusion and local channel interaction.
Here, the two methods are compared and analyzed.
In the comparison experiment, faster RCNN is used as the detector on the coco dataset, ResNet50 is the backbone, and the moments are $M_{1} + M_{2}$.
The results are reported in Table~\ref{tbl:table6}.
\begin{table}[ht]
	
	\centering
	\scalebox{0.9}{
	\begin{tabular}{ l|cccccccc }\hline
			Methods & AP & $AP_{0.5}$ & $AP_{0.75}$ & $AP_{S}$ & $AP_{M}$ & $AP_{L}$ \\ \hline
			CFC         & 36.4 & 57.6 & 39.0 & 21.2 & 39.8 & 47.0 \\
			CMC         & \textbf{38.0} & 60.0 & 40.7 & 22.4 & 42.1 & 48.2 \\
			
			\hline
	\end{tabular}}
	\caption{Effect of different moment fusion methods with ResNet-50 on COCO val2017 set.}
	\label{tbl:table6}
\end{table}

It can be seen from the experiments that the CMC method significantly improves the performance of detector, which fully proves the superiority of the CMC method.

\subsubsection{Comparison of ${M}_{triple}$ with MCA-E and MCA-S Method on the Imagenet Set}

In the Extensive Moment Aggregation (EMA) mechanism, EMA consists of base term and other terms. Although the theory shows that ${M}_{triple}$ composed of $M_{1}$, $M_{2}$, and $M_{3}$ can achieve better results, and this is also proved in the detection task. However, in the classification experiment, the performance of ${M}_{triple}$  is not as good as that of MCA-E and MCA-S.

In Table \ref{tbl:table7}, the performance of ${M}_{triple}$ ($M_{1}$ + $M_{2}$ + $M_{3}$),  MCA-E, and MCA-S  for classification task is shown. The classifier uses Resnet-50 model as the backbone. The experimental results are as follows:

\begin{table}[h]
	
	\centering
	\scalebox{0.9}{
	\begin{tabular}{ l|ll|ll }
		
		\hline
		Methods & Params & GFLOPs & Top-1 & Top-5 \\ \hline
		
		MCA-E & +6.06K & 4.127 & 76.36 & 93.14 \\ 
		MCA-S & +6.06K & 4.133  & \textbf{76.61} & \textbf{93.21} \\ 
		${M}_{triple}$ & +12.10K  & 4.139   & 76.21 & 92.99 \\ 
		\hline
	\end{tabular}
    }
	\caption{Effect of different moments with ResNet-50 on ImageNet dataset.}
	\label{tbl:table7}
\end{table}

The reason for the unsatisfactory  experimental results could be attributed to the varying ranges of values for $M_{1}$, $M_{2}$, and $M_{3}$,  which might  affect the method's performance when combined. 

%
%
%
%
%
Our research is conducted using PyTorch 1.8.2 and MindSpore 1.7.0 \cite{huawei2022deep}. The source code for this project can be accessed on GitHub at https://github.com/CSDLLab/MCA.

\section{Acknowledgments}

This work is supported by the National Key R\&D Program of China (2020YFB1313500), National Natural Science Foundation of China (T2293723, 61972347),  the Key R\&D Program of Zhejiang Province (2022C01022, 2022C01119, 2021C03003), and the Fundamental Research Funds for the Central Universities (No. 226-2022-00051). 

\bibliography{aaai24}

\end{document}